\documentclass[letterpaper]{article} 
\pdfoutput=1
\usepackage{aaai23}  
\usepackage{times}  
\usepackage{helvet}  
\usepackage{courier}  
\usepackage[hyphens]{url}  
\usepackage{graphicx} 
\usepackage{natbib}  
\usepackage{caption} 
\nocopyright

\usepackage[ruled,linesnumbered]{algorithm2e}
\usepackage{booktabs}
\usepackage{fancyvrb}
\usepackage{multirow}
\usepackage{graphicx}
\usepackage{subcaption}
\usepackage{amsmath}
\usepackage{amsfonts}

\title{PERFEX: Classifier Performance Explanations for Trustworthy AI Systems}
\author{
    Erwin Walraven\textsuperscript{\rm 1}, Ajaya Adhikari\textsuperscript{\rm 1}, Cor J. Veenman\textsuperscript{\rm 1,2}
}
\affiliations{
    \textsuperscript{\rm 1}Netherlands Organisation for Applied Scientific Research, The Hague, The Netherlands\\
    \textsuperscript{\rm 2}Leiden University, Leiden, The Netherlands
}

\newtheorem{userquestion}{User question}

\DeclareMathOperator*{\argmax}{arg\,max}

\begin{document}
\maketitle
\begin{abstract}
Explainability of a classification model is crucial when deployed in real-world decision support systems.
Explanations make predictions actionable to the user and should inform about the capabilities and limitations of the system.
Existing explanation methods, however, typically only provide explanations for individual predictions. Information about conditions under which the classifier is able to support the decision maker is not available, while for instance information about when the system is not able to differentiate classes can be very helpful.
In the development phase it can support the search for new features or combining models, and in the operational phase it supports decision makers in deciding e.g. not to use the system.
This paper presents a method to explain the qualities of a trained \emph{base} classifier, called PERFormance EXplainer (PERFEX). 
Our method consists of a \emph{meta} tree learning algorithm that is able to predict and explain under which conditions the base classifier has a high or low \emph{error} or any other classification performance metric.
We evaluate PERFEX using several classifiers and datasets, including a case study with urban mobility data. 
It turns out that PERFEX typically has high meta prediction performance even if the base classifier is hardly able to differentiate classes, while giving compact performance explanations.
\end{abstract}

\section{Introduction}
Decision support systems based on machine learning models are being developed for a growing number of domains. To deploy these systems for operational use, it is crucial that the system provides tangible explanations.
It needs to be transparent about the underlying inference process of its predictions and about its own limitations.
For example, in a medical setting it is important that a doctor knows under which circumstances the system is unable to provide reliable advice regarding a diagnosis~\cite{Papanastasopoulos2020}.
Similarly, when a decision support system is used for investment decisions in policy making, then it is important that the users of the system get informed about the uncertainty associated with the advice it provides~\cite{Arroyo2019}.
The importance of explanation capabilities is also emphasized by the guidelines on trustworthy AI from the European Commission~\cite{EU2019}, which includes explainability about capabilities and limitations of AI models as a key requirement.

Developing methods for explainability in machine learning has gained significant interest in recent years~\cite{Burkart2021}. 
Global explanations give an overall view of the knowledge encoded in the model. This is very relevant for knowledge discovery like in biology or medical applications. 
Examples are model coefficients as in logistic regression \cite{Cox1958} and indications of feature importance such as with random forests~\cite{Breiman2002} and gradient boosting~\cite{Chen2016}. Local explanations on the other hand explain predictions for individual datapoints. For example, SHAP and LIME explain the class prediction of a datapoint by highlighting features which are locally most influencing the prediction~\cite{Lundberg2017,Ribeiro2016}.

In addition to explaining class predictions, methods are needed that explain the performance of a classifier. 
Such explanations can be used by a data scientist to understand under which circumstances a base classifier does or does not not perform well.
If the explanation defines that the model does not perform well for a specific subset of the data, then the data scientist may decide to look for additional data, additional features, or otherwise attempt to improve the model in a focused way.
The explanations can also be used to inform e.g. a consultant or medical doctor about circumstances in which a model cannot be trusted, which is also relevant for engineers who bring models to production.
In existing literature only a method for explaining the uncertainty of individual predictions has been proposed (e.g., \citeauthor{Antoran2021}, \citeyear{Antoran2021}). For explaining the performance characteristics and limitations of classifiers globally, no methods have been published to the best of our knowledge.

This paper presents a model-agnostic PERFormance EXplainer (PERFEX) to derive explanations about characteristics of classification models.
Given a \emph{base} classifier, a dataset and a classification performance metric such as the prediction accuracy, we 
propose a \emph{meta} learning algorithm that separates the feature space in regions with high and low prediction accuracy and enables to generate compact explanations for these regions.
In the following sections we define the problem formally and overview related work. Then, we describe PERFEX in detail. We evaluate the method in experiments based on several classification methods and datasets including our own case study.
The experiments show that PERFEX provides clear explanations in scenarios where explanations from SHAP and LIME are not sufficient to gain trust.
We finalize the paper with our conclusions. 

\section{Problem Statement}
We consider a classification task in which a base classifier~$\mathcal{C}$ is trained to assign a datapoint~$x$ to a class.
The set~$\mathcal{K}=\{c_1, c_2, \ldots, c_k \}$ contains all~$k$ classes considered, and~$\mathcal{C}(x) \in \mathcal{K}$ denotes the class to which datapoint~$x$ belongs according to~$\mathcal{C}$.
The classifier is trained using a tabular dataset~$\mathcal{X}_t$ containing~$n$ datapoints, and for each datapoint~$x_i \in \mathcal{X}_t$ the true class label is denoted by~$y_i \in \mathcal{K}$.
Each datapoint in the dataset is defined by~$m$ feature values, and we use~$x^j$ to refer to feature value~$j$ of datapoint~$x$.

The prediction performance of classifier~$\mathcal{C}$ can be measured using standard metrics, such as 
accuracy, precision, recall, F1-score and expected calibration error~\cite{Guo2017}.
We define the prediction performance metric~$\mathcal{M}$ as a function that takes a classifier~$\mathcal{C}$, test datapoints~$x_1,\ldots, x_p$ and true labels~$y_1, \ldots, y_p$ as input, and it computes a real-valued score as output.
The problem we consider is: given a classifier~$\mathcal{C}$, a metric~$\mathcal{M}$, an independent dataset~$\mathcal{X}$ and corresponding ground truth labels~$\mathcal{Y}$, find a compact explanation for subgroups of the data having either low or high (sub) performance.
The compactness refers to the amount of information that the explanation presents to the user.

\begin{figure}[t]
\includegraphics[width=\linewidth]{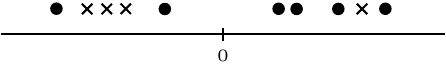}
\caption{\label{fig:example}One-dimensional dataset with correct predictions (dot) and incorrect predictions (cross)}
\end{figure}

As an example we consider prediction accuracy as performance metric~$\mathcal{M}$, and we visually illustrate the problem based on a one-dimensional dataset with feature~$z$, as shown in Figure~\ref{fig:example}. The symbols indicate whether predictions from a given base classifier are correct (dot) or not (cross) when predicting for the ten datapoints that are shown. The overall prediction accuracy is~$0.6$. However, this number does not tell us under which circumstances the classifier performs well, and when it does not perform well. We would like to create explanations which tell that the classifier does not perform well for~$z<0$~(accuracy~$2/5=0.4$), while it does perform well otherwise~(accuracy~$4/5=0.8$).
Instead of accuracy other performance metrics~$\mathcal{M}$ may be used, such as precision, recall, F1-score and expected calibration error.

\section{Related Work}
In this section we overview the work related to the stated problem ranging from model agnostic individual prediction to cluster based explanations and explaining uncertainties. 

First, SHAP~\cite{Lundberg2017} and LIME~\cite{Ribeiro2016} can be used to create explanations about individual predictions. 
SP-LIME~\cite{Ribeiro2016} is a variant of LIME which aims to enable a user to assess whether a model can be trusted by providing an explanation for a group of samples as set of individual explanations.
This is problematic in domains with many features, it requires that the user inspects many instances, and it is unclear whether a set of local explanations gives global understanding of a model.
Anchors~\cite{Ribeiro2018} is related to LIME and aims to explain how a model behaves on unseen instances, but only locally.
K-LIME is another variant of LIME, which is part of the H2O Driverless AI platform~\cite{H2O}.
It performs a $k$-means clustering, and for each cluster it fits a linear model to explain features influencing the predictions in that cluster. 
In contrast to our problem, it uses a normal classification model fitting criterion instead of explaining a (base) learner using its performance metric.

Interpretable clustering~\cite{Bertsimas2021} clusters data based on a tree structure.
It derives an optimal clustering tree using mixed-integer optimization, and the branches in the tree structure make the clustering interpretable.
Although this approach may deliver compact cluster explanations, like the LIME variants, it models the distribution of the data itself instead of the prediction structure of a base learner. 
A clustering based on the predictions of a model cannot be easily integrated in this exact optimization framework, especially if the computation of the performance metric is non-linear. 

Explanations of prediction characteristics of a classifier is related to explanations of uncertainty.
The CLUE method~\cite{Antoran2021} can be used to explain which parts of the input of a deep neural network cause uncertainty by providing a counterfactual explanation in the input space.
CLUE only provides an uncertainty explanation for an individual input, and it cannot be used to inform the user about the circumstances under which a model is uncertain.
Our work also relates to Interpretable Confidence Measures~(ICM), which uses the accuracy as a proxy for uncertainty~\cite{VanderWaa2020}.
A prediction for a datapoint is considered to be uncertain if the classifier makes mistakes for similar datapoints.
Our problem is to provide e.g. uncertainty explanations for groups of datapoints, whereas ICM only focuses on individual datapoints.

Finally, there is a link with Emerging Pattern Mining~(EPM), which can be used to capture contrasts between classes~\cite{Dong1999}.
An important difference is that EPM aims at find patterns in data, while we aim at finding patterns in the modeled data (by potentially any classifier).

\section{Classifier PERFormance EXplainer}
\label{sec:trees}
This section describes our method to find compact explanations for subsets of datapoints based on local high or low performance of the base learner. As overviewed in the Related Work section applying clustering algorithms, such as $k$-means, is not suitable because $k$-means does not cluster based on~$\mathcal{M}$. 
A clustering based on a decision tree can address this problem, because datapoints in leafs can be seen as clusters and the branch conditions in the tree can be used to extract explanations.
If the classifier accuracy is used as metric~$\mathcal{M}$, then a standard decision tree can be fitted which uses train targets which equal 1 if the base classifier predicts correctly for a datapoint, and 0 otherwise. 
This would yield a tree which distinguishes subsets of data with low accuracy from subsets of data with high accuracy, and allows for explanations.
However, for other performance metrics~$\mathcal{M}$ such targets cannot be defined.
We introduce PERFEX, model-agnostic method to explain the prediction performance of a base classifier for any performance metric~$\mathcal{M}$.

\subsection{Creating Subsets of Data using Tree Structure}
\label{sec:treegeneration}
\begin{figure}[t]
\includegraphics[width=\linewidth]{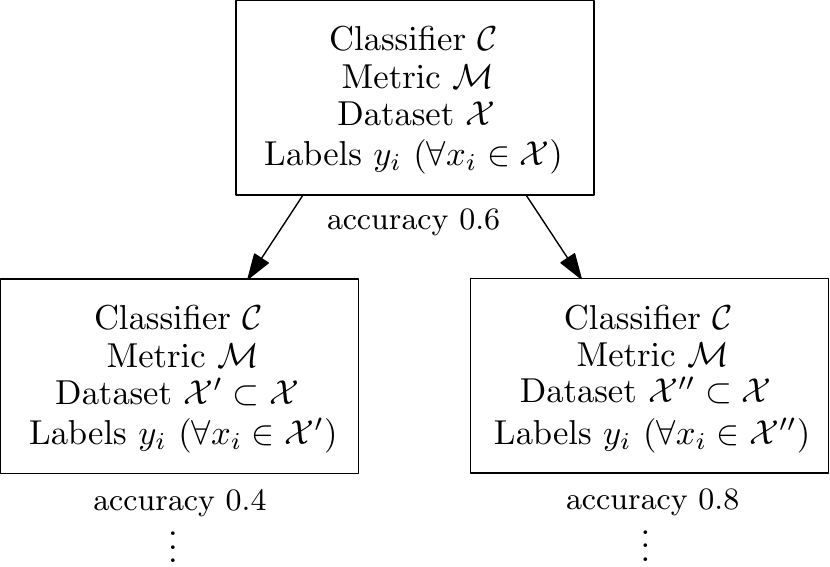}
\caption{\label{fig:decomposition}Splitting dataset~$\mathcal{X}$ into subsets~$\mathcal{X}'$ and~$\mathcal{X}''$}
\end{figure}

The basic idea of PERFEX is to divide~$\mathcal{X}$ up in a hierarchical manner leading to a tree-structured meta learner.
This enables us to naturally split the data based on a split condition that depends on~$\mathcal{M}$, similar to the construction of classification trees.
More importantly, through the hierarchical process the tree has typically a limited depth, such that when we use the branches as conditions in a decision rule, it leads to a compact explanation.
The process is schematically illustrated in Figure~\ref{fig:decomposition}.
For convenience we illustrate the tree construction based on the same accuracy values as in Figure~\ref{fig:example}.
In the root node we consider a classifier~$\mathcal{C}$, metric~$\mathcal{M}$, dataset~$\mathcal{X}$ and the corresponding labels.
The prediction metric score for $\mathcal{X}$ can be obtained by evaluating~$\mathcal{M}$, which gives accuracy 0.6 in the figure.
This value has been computed using the full dataset~$\mathcal{X}$, but it does not enable the user to understand when this metric value is low or high.
We provide this additional understanding to the user by decomposing~$\mathcal{X}$ into two subsets~$\mathcal{X}' \subset \mathcal{X}$ and $\mathcal{X}'' \subset \mathcal{X}$, such that $\mathcal{M}$ evaluates to a low value for $\mathcal{X}'$ and to a high value for $\mathcal{X}''$.
This process is illustrated by the child nodes, which evaluate to an accuracy of 0.4 and 0.8, respectively.
The branch conditions in the tree can be used to explain to a user when the performance metric evaluates to a low or high value.
\begin{algorithm}[t]
\SetKwInOut{Input}{input}
\SetKwInOut{Output}{output}
\Input{classifier~$\mathcal{C}$, dataset~$\mathcal{X}$, labels~$y_i$ ($\forall x_i \in \mathcal{X}$), prediction metric~$\mathcal{M}$, minimum subset size~$\alpha$}
\Output{subsets~$\mathcal{X}' \subset \mathcal{X}$ and $\mathcal{X''} \subset \mathcal{X}$ with corresponding labels, split condition~$s$}
$\mathcal{X}' \leftarrow \emptyset$,~~$\mathcal{X''} \leftarrow \emptyset$,~~$s\leftarrow (0,0)$, ~~$\beta \leftarrow 0$\\
\For{$j=1,\ldots,m$}{
  \ForEach{unique value $v$ of feature~$j$ in $\mathcal{X}$}{
    $\hat{\mathcal{X}}' \leftarrow \emptyset$,~~$\hat{\mathcal{X}}'' \leftarrow \emptyset$\label{line:subsetstart}\\
    \ForEach{$x \in \mathcal{X}$}{
      \uIf{$x^j \leq v$}{\label{line:comparison}
        $\hat{\mathcal{X}}' \leftarrow \hat{\mathcal{X}}' \cup \{ x \}$
      }
      \Else{
        $\hat{\mathcal{X}}'' \leftarrow \hat{\mathcal{X}}'' \cup \{ x \}$
      }
    }\label{line:subsetend}
    $e' \leftarrow$ evaluate $\mathcal{M}$ using $\mathcal{C}$, $\hat{\mathcal{X}}'$ and labels\label{line:beststart}\\
    $e'' \leftarrow$ evaluate $\mathcal{M}$ using $\mathcal{C}$, $\hat{\mathcal{X}}''$ and labels\\
    $\beta' \leftarrow |e' - e''|$\\
    \If{$\beta' > \beta$ and $|\hat{\mathcal{X}}'| \geq \alpha$ and $|\hat{\mathcal{X}}''| \geq \alpha$}{
      $\beta \leftarrow \beta'$,~$\mathcal{X}' \leftarrow \hat{\mathcal{X}}'$,~$\mathcal{X}'' \leftarrow \hat{\mathcal{X}}''$,~$s \leftarrow (j, \beta)$
    }\label{line:bestend}
  }
}
\caption{\label{alg:treegeneration}PERFEX}
\end{algorithm}

The tree structure in Figure~\ref{fig:decomposition} can be automatically created using an algorithm that closely resembles to procedure for generating decision trees for classification and regression~\cite{Breiman1984}.
A key difference is that we use a split condition based on~$\mathcal{M}$ during the tree generation procedure, rather than using e.g. the Gini impurity.
Algorithm~\ref{alg:treegeneration} shows how to split a dataset~$\mathcal{X}$ for all possible features into subsets~$\mathcal{X}' \subset \mathcal{X}$ and $\mathcal{X}'' \subset \mathcal{X}$ using prediction metric~$\mathcal{M}$ as split criterion.
It enumerates all possible splits into subsets $\mathcal{X}'$ and $\mathcal{X}''$.
For numerical and binary features a less-than-or-equal condition can be used on line~\ref{line:comparison}, and for categorical features an equality condition should be used.
For features with continuous values, it may be practical to consider only a fixed number of quantiles, rather than enumerating all unique values. 
After creating subsets on lines~\ref{line:subsetstart}-\ref{line:subsetend}, it uses~$\mathcal{M}$ to evaluate the metric value for both subsets, and it keeps track of the best split found so far.
The quality of a split is determined by computing the difference between the performance metric values of both subsets.
Since we want to distinguish subsets with low and high metric values, the algorithm returns the subsets with maximum difference.
The split condition corresponding to the best split is stored in the tuple~$s$, which contains both the index of the feature and the feature value used for splitting.
Algorithm~\ref{alg:treegeneration} shows how one node of the tree ($\mathcal{X}$) is divided into two child nodes ($\mathcal{X}'$, $\mathcal{X}''$).
In order to create a full tree, the algorithm should be applied again to~$\mathcal{X}'$ and~$\mathcal{X}''$.
This process repeats until a fixed depth is reached.
Another stop criterion based on confidence intervals is discussed below.

\subsection{Confidence Intervals on Values of~$\mathcal{M}$}
Splitting data using Algorithm~\ref{alg:treegeneration} should terminate if the size of either~$\mathcal{X}'$ or $\mathcal{X}''$ becomes too small to provide a good estimation of metric~$\mathcal{M}$.
This can be assessed based on a confidence interval on~$e'$ and $e''$.
We only discuss the derivation for~$e'$ because for~$e''$ the procedure is identical.
The actual derivation is dependent on the chosen metric~$\mathcal{M}$.
Below we illustrate it for the metrics accuracy and precision.

For accuracy the estimator~$e' = u~/~|\mathcal{X}'|$ can be used, in which~$u$ represents the total number of correct predictions.
The estimator~$e'$ follows a binomial distribution, and therefore we can use a binomial proportion confidence interval:
\begin{align}
    \left( e' - z \sqrt{\frac{e'(1-e')}{|\mathcal{X}'|}}, e' + z \sqrt{\frac{e'(1-e')}{|\mathcal{X}'|}} \right),
\end{align}
in which~$z$ denotes the Z-score of the desired confidence level~\cite{Pan2002}.
Given maximum interval width~$D$, combined with the insight that the term $e'(1-e')$ takes a value that is at most~$0.25$, we obtain the minimum number of datapoints, which can be used as termination condition:
\begin{align}
z \sqrt{\frac{0.25}{|\mathcal{X}'|}} = \frac{D}{2}~~~~\Rightarrow~~~~|\mathcal{X}'| = \frac{z^2}{D^2}.
\end{align}
For example, when using a 95 percent confidence level and maximum interval width~$0.1$, the minimum number of datapoints in $\mathcal{X}'$ equals $1.96^2~/~0.1^2 \approx 384$.

For other proportion metrics such as precision the derivation is slightly different, because precision does not depend on all datapoints in~$\mathcal{X}'$.
For example, the precision for class~$c_i$ equals~$u~/~|\{ x \in \mathcal{X}'~|~ \mathcal{C}(x)=c_i \} |$, in which $u$ denotes the number of datapoints in $\{ x \in \mathcal{X}'~|~ \mathcal{C}(x)=c_i \}$ which were predicted correctly.
By applying the same derivation as above, it can be seen that the termination condition for tree generation should be based on the number of datapoints in $\{ x \in \mathcal{X}'~|~ \mathcal{C}(x)=c_i \}$ rather than~$\mathcal{X}'$.

\subsection{Tree Evaluation using Test Set}
\begin{algorithm}[t]
\SetKwInOut{Input}{input}
\SetKwInOut{Output}{output}
\Input{tree~$\mathcal{T}$ created by recursively applying Algorithm~\ref{alg:treegeneration}, metric~$\mathcal{M}$, classifier~$\mathcal{C}$, dataset~$\mathcal{X}$, test set $\bar{\mathcal{X}}$, and labels}
\Output{mean absolute error~$\hat{e}$, metric difference~$d$}
$L \leftarrow$~set of leafs in~$\mathcal{T}$,~~$\hat{e} \leftarrow 0$\\
\ForEach{leaf~$l \in L$}{
$\mathcal{X}_l \leftarrow$~datapoints in leaf~$l$ after applying~$\mathcal{T}$ to $\mathcal{X}$\\
$\bar{\mathcal{X}}_l \leftarrow$~datapoints in leaf~$l$ after applying~$\mathcal{T}$ to $\bar{\mathcal{X}}$\\
$e_l \leftarrow$~evaluate $\mathcal{M}$ using $\mathcal{C}$, $\mathcal{X}_l$ and labels\\
$\bar{e}_l \leftarrow$~evaluate $\mathcal{M}$ using $\mathcal{C}$, $\bar{\mathcal{X}}_l$ and labels\\
$\hat{e} \leftarrow \hat{e} + |e_l-\bar{e}_l|$
}
$\hat{e} \leftarrow \hat{e}~/~|L|$,~~~~$d \leftarrow (\max_{l \in L} e_l - \min_{l \in L} e_l)$
\caption{\label{alg:error}Estimate quality of tree using test set}
\end{algorithm}

The tree quality can also be evaluated using a separate test set~$\bar{\mathcal{X}}$.
First, the datapoints in $\bar{\mathcal{X}}$ are assigned to leafs.
After that, the metric value in each leaf can be computed based on the assigned datapoints.
Intuitively, it can be expected that the metric value for datapoints in a leaf of the tree is similar for $\mathcal{X}$ and $\bar{\mathcal{X}}$, regardless of the performance of the base classifier and regardless of the performance metric~$\mathcal{M}$.
For example, if the accuracy in all the leafs of the tree is low, then the estimated accuracy in the leafs will also be low when using another dataset from the same distribution.
Algorithm~\ref{alg:error} shows how the tree quality is determined by computing the mean absolute error based on the errors of the individual leafs.
The output variable~$d$ can be used to assess to what extent PERFEX distinguishes subsets with low and high metric values.

\subsection{Generating Explanations}
\label{sec:generate_explanations}
The tree structure created by Algorithm~\ref{alg:treegeneration} can be used to extract explanations that can be presented to a user in a text-based format.
Each leaf in the tree represents a subset of the data with a corresponding metric value.
Therefore, we can print information about the leaf which explains to the user how the subset of data has been constructed, and what the metric value is, as illustrated below for one leaf:
\begin{Verbatim}[fontsize=\small]
There are 134 datapoints for which the
following conditions hold:
  length > 10.77, length <= 12.39
and for these datapoints accuracy is 0.68
\end{Verbatim}
For each leaf we print the number of datapoints, the prediction metric value computed on the same subset of data, and the conditions that were used to split the data.
The conditions can be extracted from the tree by taking the conditions used in the nodes along the path from root to leaf.

\subsection{Example using 2D dataset}
\begin{figure}[t]
\centering
\includegraphics[width=0.9\linewidth]{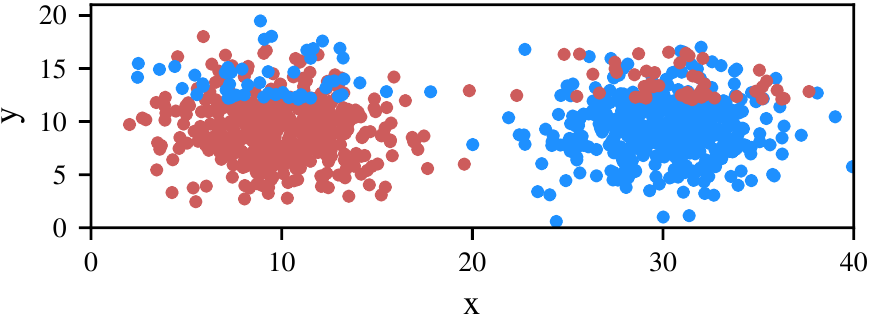}
\caption{\label{fig:exampledataset}Example dataset with two features and two classes}
\end{figure}

We provide an example using a dataset with two features, which shows visually how PERFEX creates an explanation.
It will also show that explanations for the class prediction are not the same as the explanations based on~$\mathcal{M}$.
The dataset is shown in Figure~\ref{fig:exampledataset} and consists of the classes red and blue, generated using Gaussian blobs with centers~$(10,10)$ and~$(30,10)$.
For the purpose of the example we flip labels of some datapoints for which~$y>12$.
The majority of the datapoints for which~$x < 20$ belongs to red, and the majority belongs to blue if~$x \geq 20$.
Our base classifier is a decision tree with depth~1, predicting red if~$x < 20$ and blue otherwise.
We investigate when the base classifier has a low accuracy by applying Algorithm~\ref{alg:treegeneration} with accuracy as metric~$\mathcal{M}$:
\begin{Verbatim}[fontsize=\footnotesize]
There are 100 datapoints for which the
following conditions hold:
 y > 10.96
and for these datapoints accuracy is 0.72
\end{Verbatim}
\begin{Verbatim}[fontsize=\footnotesize]
There are 200 datapoints for which the
following conditions hold:
 y <= 10.96
and for these datapoints accuracy is 1.0
\end{Verbatim}

This explanation shows that the accuracy is lower if~$y > 10.96$, which is also the area in which datapoints belong to two classes.
More importantly, it shows that the explanation for accuracy depends on~$y$, whereas the prediction made by the base classifier (and its explanation) only depend on~$x$.

\section{Experiments}
We present the results of our experiments based on Gaussian data as well as several standard classifiers and datasets.

\subsection{Evaluation of Tree Error with Gaussian Data}
\begin{figure}[t]
\centering
\includegraphics[width=0.87\linewidth]{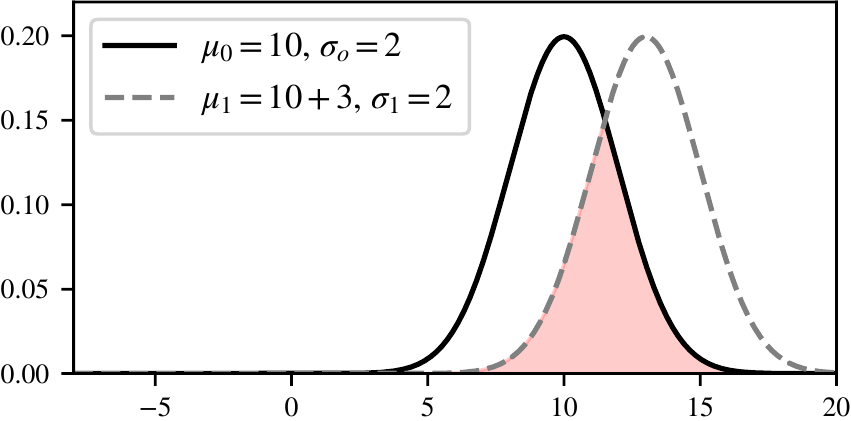}
\caption{\label{fig:bayes_error}Distribution with two classes defined by two Gaussians and one feature. The shaded area is the region of error.}
\end{figure}

\begin{figure}[t]
\centering
\begin{subfigure}{0.45\linewidth}
  \centering
  \includegraphics[width=0.93\linewidth]{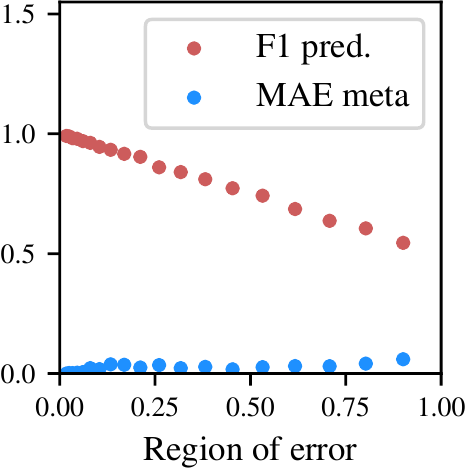}
  \caption{Error for increasing region of error}
  \label{fig:results_gaussians}
\end{subfigure}%
\hfill
\begin{subfigure}{0.45\linewidth}
  \centering
  \includegraphics[width=\linewidth]{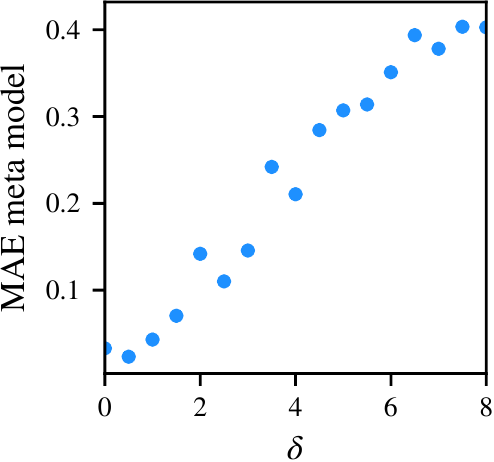}
  \caption{Error of meta model for shifted data distribution}
  \label{fig:results_distribution_shift}
\end{subfigure}
\caption{Results of experiments with Gaussian data}
\end{figure}

We start with two experiments to empirically study two hypotheses from the previous section.
We use data from Gaussian distributions, allowing us to carefully control the difficulty of the prediction task.
In our first experiment we show that PERFEX can be used to model a chosen prediction metric even if the original class prediction task is hard.
We assume that the data comes from a one-dimensional dataset defined by two Gaussians, as shown in Figure~\ref{fig:bayes_error}.
The Gaussian with~$\mu=10$ and~$\sigma=2$ corresponds to the first class, and remains fixed.
The datapoints of the second class follow a Gaussian distribution with~$\mu=10+\delta$ and $\sigma=2$.
The parameter~$\delta>0$ is used to control the overlap of both Gaussians, which affects the difficulty of the prediction task.
In the figure this is visualized for~$\delta=3$.
Since the standard deviation of both distributions is the same, the difficulty of the prediction task can be expressed using the region of error, which is visualized using the shaded red area.
We define a classifier which predicts the class for a datapoint~$x$ by taking the class for which the probability density is maximum:~$\mathcal{C}(x) = \argmax_{i \in \{0,1\}} f(\mu_i, \sigma_i, x)$, in which~$f$ denotes the probability density function.
It can be expected that the prediction performance of the classifier drops if the region of error grows.
This is confirmed in Figure~\ref{fig:results_gaussians}, which shows the weighted F1 score of the classifier for an increasing region of error.
We also created a PERFEX tree for accuracy, for which the mean absolute error (MAE, computed by Algorithm~\ref{alg:error}) is also shown.
The error is close to zero, which confirms that PERFEX can model the accuracy of the classifier~$\mathcal{C}$ even if the performance of this classifier is low.

Now we show that a generated tree can be used to model a prediction metric for a given classifier if the data used for creating the meta decision tree comes from the same distribution as the data used for creating the base classifier.
We conduct an experiment in which we measure the error of the PERFEX tree, and we gradually shift the data distribution for creating the tree, which causes the error to increase.
We use two Gaussians for creating the prediction model~$\mathcal{C}$, with~$\mu_0 = 10$, $\mu_1 = 13$ and~$\sigma_0 = \sigma_1 = 2$.
The data used for creating the PERFEX tree uses the same distributions, except that $\mu_1 = 13 + \delta$ with $\delta \geq 0$.
If~$\delta = 0$, then all datasets come from the same distribution, and in that case the error of the meta decision tree is low, as can be seen in Figure~\ref{fig:results_distribution_shift}.
If we shift the distribution of the data for creating our tree by setting~$\delta > 0$, then we expect that the error increases due to this mismatch in the data.
The figure confirms this, and it shows that a tree can be fitted only if its training data comes from the same distribution as the classifier training data.

\subsection{Evaluation on Several Datasets and Models}
We apply PERFEX to different datasets, classifiers and split conditions based on several metrics.
Given that the meta tree needs sufficient data to create generalizable clusters, 4 classification datasets with at least 1000 datapoints from the UCI repository \cite{uci:2019} were chosen: abalone, car evaluation, contraceptive method choice, and occupancy detection.
While experimenting, we noticed that the classification of occupancy had almost perfect scores on the test-set.
In that case, the meta model would not be able to create clusters.
For that reason, we made the classification task more difficult by only including two features in the dataset: CO2 and temperature.
Finally, we also included a fifth 2D dataset called \emph{gaussian blobs}, which contains three clusters of datapoints that are partially overlapping.
These clusters were sampled from a isotropic Gaussian distribution with cluster centers (10, 10), (20,12) and (15, 15), and a standard deviation of 3.
We use this dataset to validate whether the tree is able to distinguish the non-overlapping regions with perfect scores and the overlapping regions with lower scores.

\begin{table*}[t]
\small
\centering
\def\arraystretch{0.1}
\begin{tabular}{llrrrrrrr}
\toprule
\textbf{Dataset}            & $\mathcal{C}$ & \textbf{Accuracy of} $\mathcal{C}$ & \textbf{Num. leafs} & \textbf{Tree depth} & \textbf{Min accuracy} & \textbf{Max accuracy} & \textbf{MAE} & \textbf{STD AE} \\
\midrule
\multirow{5}{*}{Abalone}    & SVC               & 53,6\%                                   & 6                                                                                            & 6                                  & 32,1\%                                                                                         & 84,3\%                                                                                         & 5,7\%                            & 4,6\%                                \\
                            & LR                & 56,5\%                                   & 6                                                                                            & 6                                  & 37,2\%                                                                                         & 83,5\%                                                                                         & 4,9\%                            & 3,8\%                                \\
                            & RF                & 54,5\%                                   & 6                                                                                            & 6                                  & 43,9\%                                                                                         & 82,7\%                                                                                         & 3,6\%                            & 2,7\%                                \\
                            & DT                & 52,5\%                                   & 6                                                                                            & 6                                  & 37,0\%                                                                                         & 78,7\%                                                                                         & 5,1\%                            & 3,0\%                                \\
                            & KNN               & 49,0\%                                   & 6                                                                                            & 6                                  & 32,2\%                                                                                         & 82,5\%                                                                                         & 4,2\%                            & 2,6\%                                \\
\midrule
       & SVC               & 95,0\%                                   & 2                                                                                            & 2                                  & 87,6\%                                                                                         & 99,1\%                                                                                         & 1,1\%                            & 0,7\%                                \\
Car                            & LR                & 90,5\%                                   & 4                                                                                            & 4                                  & 74,8\%                                                                                         & 100,0\%                                                                                        & 1,9\%                            & 2,9\%                                \\
evaluation                            & RF                & 95,2\%                                   & 2                                                                                            & 2                                  & 89,2\%                                                                                         & 98,5\%                                                                                         & 1,4\%                            & 0,2\%                                \\
                            & DT                & 94,8\%                                   & 3                                                                                            & 3                                  & 85,8\%                                                                                         & 100,0\%                                                                                        & 1,6\%                            & 0,9\%                                \\
                            & KNN               & 87,5\%                                   & 4                                                                                            & 4                                  & 65,5\%                                                                                         & 99,4\%                                                                                         & 3,8\%                            & 3,9\%                                \\
\midrule
    & SVC               & 44,3\%                                   & 3                                                                                            & 3                                  & 33,0\%                                                                                         & 63,3\%                                                                                         & 7,2\%                            & 5,3\%                                \\
Contraceptive                            & LR                & 53,4\%                                   & 3                                                                                            & 3                                  & 41,1\%                                                                                         & 63,8\%                                                                                         & 8,8\%                            & 2,0\%                                \\
method                            & RF                & 51,4\%                                   & 4                                                                                            & 4                                  & 38,3\%                                                                                         & 62,4\%                                                                                         & 4,3\%                            & 2,4\%                                \\
choice                            & DT                & 52,3\%                                   & 3                                                                                            & 3                                  & 38,7\%                                                                                         & 62,6\%                                                                                         & 4,7\%                            & 1,3\%                                \\
                            & KNN               & 48,2\%                                   & 4                                                                                            & 3                                  & 36,0\%                                                                                         & 62,4\%                                                                                         & 9,5\%                            & 6,2\%                                \\
\midrule
  & SVC               & 86,8\%                                   & 12                                                                                           & 6                                  & 20,6\%                                                                                         & 98,8\%                                                                                         & 3,6\%                            & 2,4\%                                \\
 Occupancy                           & LR                & 81,9\%                                   & 10                                                                                           & 6                                  & 21,1\%                                                                                         & 98,5\%                                                                                         & 3,9\%                            & 3,6\%                                \\
detection                            & RF                & 94,5\%                                   & 7                                                                                            & 6                                  & 80,3\%                                                                                         & 100,0\%                                                                                        & 2,0\%                            & 1,7\%                                \\
                            & DT                & 93,5\%                                   & 9                                                                                            & 6                                  & 72,5\%                                                                                         & 98,3\%                                                                                         & 3,5\%                            & 2,0\%                                \\
                            & KNN               & 88,6\%                                   & 9                                                                                            & 6                                  & 66,0\%                                                                                         & 98,6\%                                                                                         & 3,7\%                            & 2,0\%                                \\
\midrule
 & SVC               & 80,3\%                                   & 8                                                                                            & 6                                  & 73,4\%                                                                                         & 100,0\%                                                                                        & 2,0\%                            & 2,8\%                                \\
Gaussian                            & LR                & 80,2\%                                   & 8                                                                                            & 6                                  & 73,1\%                                                                                         & 100,0\%                                                                                        & 1,2\%                            & 1,2\%                                \\
blobs                            & RF                & 78,6\%                                   & 9                                                                                            & 6                                  & 69,9\%                                                                                         & 100,0\%                                                                                        & 2,3\%                            & 1,5\%                                \\
                            & DT                & 76,9\%                                   & 7                                                                                            & 6                                  & 70,1\%                                                                                         & 98,5\%                                                                                         & 2,8\%                            & 2,8\%                                \\
                            & KNN               & 75,6\%                                   & 8                                                                                            & 6                                  & 67,7\%                                                                                         & 100,0\%                                                                                        & 3,5\%                            & 2,6\%                               \\
\bottomrule
\end{tabular}
\caption{\label{table:experiment_different_datasets_models}Evaluation of PERFEX using several datasets and base classifiers with accuracy as metric~$\mathcal{M}$}
\end{table*}

Each dataset was split into a train set (50\%), test 1 set (25\%) and test 2 set (25\%), in a stratified manner according to the target.
The train set was used to train 5 base classifiers: Logistic Regression (LR), Support Vector Machine with RBF kernel (SVM), Random Forest (RF), Decision Tree (DT), and KNN with K=3.
Test set 1 was used to evaluate the base classifier and to build the PERFEX tree with maximum depth 6.
The tree is used to cluster the datapoints of test set 1 and test set 2, separately.
This tree is evaluated by comparing the accuracy scores of the corresponding clusters of the test sets using Mean Absolute Error (MAE), as described in Algorithm~\ref{alg:error}.
This shows whether PERFEX is able to generalize the accuracy estimates to an unseen dataset.

Table \ref{table:experiment_different_datasets_models} shows both the performance of the base classifiers and the corresponding PERFEX tree based on accuracy.
The classification models have a diverse accuracy, ranging from 51\% to 95\%.
For PERFEX the table shows the amount of leaves, the depth of the tree, the minimum and maximum accuracies among the leaves, the MAE, and the STD of the Absolute Error (AE).
PERFEX is able to separate datapoints with high and low accuracy, with the lowest difference of 9.3\% (Car RF) and the highest of 78.2\% (Contraceptives SVC).
For the gaussian blobs, we see clusters with perfect or almost perfect scores, as expected.
The MAE is generally low, ranging from 1.1\% to 9.5\%.
We also see a pattern in which classification models with high accuracy result in lower MAE.
The supplement contains results for other metrics~$\mathcal{M}$, and more details on the datasets and our code\footnote{\url{https://github.com/erwinwalraven/perfex}}.

\subsection{Limitations of SHAP and LIME}
SHAP and LIME were introduced as methods to explain why a classifier makes a prediction, and to gain trust about the prediction.
However, this can be dangerous in practice because SHAP and LIME provide explanations regardless of the classifier performance.
We show that circumstances exist in which SHAP and LIME mark specific features as very important for a high-confidence prediction, while PERFEX clearly indicates that people should not rely on the classifier.

\begin{figure*}[t]
\centering
\begin{subfigure}{0.24\linewidth}
  \centering
  \includegraphics[height=2.8cm]{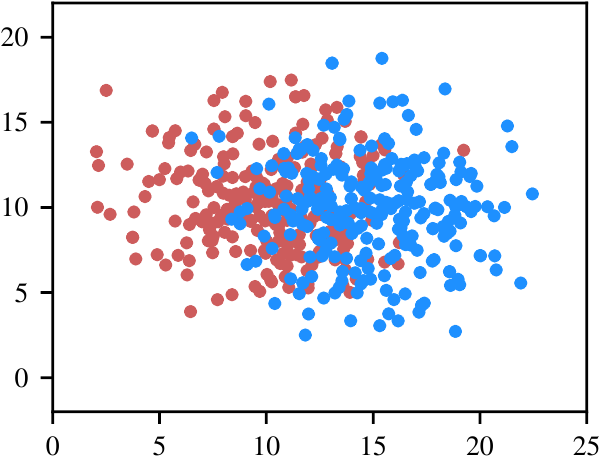}
  \caption{Dataset}
  \label{fig:lime_shap_dataset}
\end{subfigure}%
\hfill
\begin{subfigure}{0.24\linewidth}
  \centering
  \includegraphics[height=2.8cm]{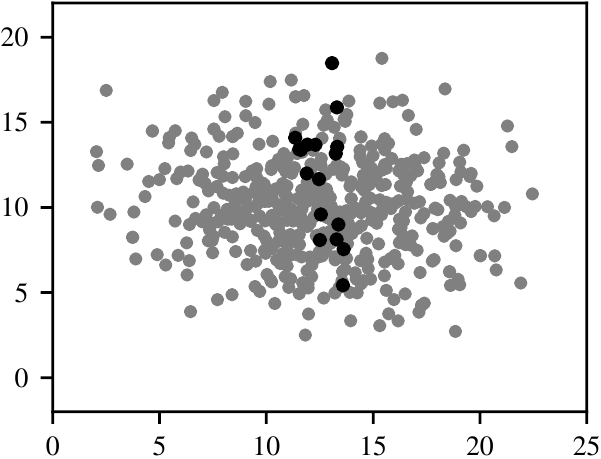}
  \caption{Highlighted datapoints}
  \label{fig:lime_shap_highlight}
\end{subfigure}%
\hfill
\begin{subfigure}{0.24\linewidth}
  \centering
  \includegraphics[height=2.8cm]{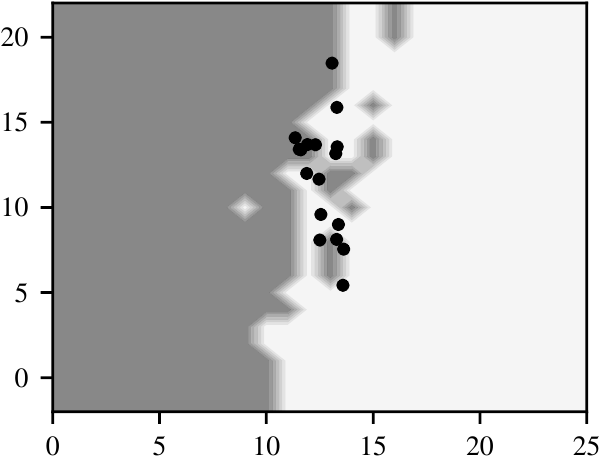}
  \caption{Decision boundary}
  \label{fig:lime_shap_decision_boundary}
\end{subfigure}%
\hfill
\begin{subfigure}{0.24\linewidth}
  \centering
  \includegraphics[height=2.8cm]{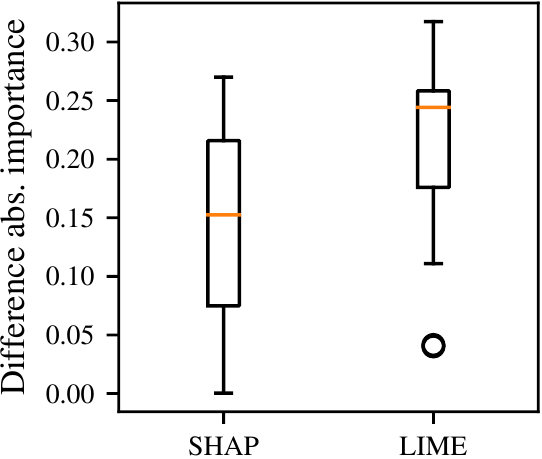}
  \caption{Feature importance}
  \label{fig:lime_shap_boxplot}
\end{subfigure}%
\caption{Results of experiment with SHAP and LIME, with feature $x^0$ horizontal and feature $x^1$ vertical}
\end{figure*}

We consider a scenario in which a doctor uses a classifier to create predictions for patients that arrive, and SHAP and LIME are used to inform the doctor about the importance of features.
The classifier is a random forest that was trained by a data scientist using the dataset shown in Figure~\ref{fig:lime_shap_dataset}.
It can be seen that it may be difficult to predict in the area where both classes are overlapping.
However, the doctor is not aware of this, and during model development the data scientist concluded based on accuracy (0.76) that the performance of the classifier is sufficient.
Suppose that a patient arrives with features~$x^0=10$ and~$x^1=12$.
It may happen that the classifier assigns a high score to one class, and SHAP and LIME highlight one feature as much more important than the other.
This is not desirable, because the doctor gets the impression that the system can be trusted, while the classifier should not be used for such patients.

We now show that datapoints exist for which the described problem arises.
According to PERFEX the cluster with the lowest accuracy (0.51) is defined by~$11.2 \leq x^0 \leq 13.6$.
In Figure~\ref{fig:lime_shap_highlight} we highlight datapoints that belong to this cluster, and for which two additional properties hold.
First, the random forest assigns at least score 0.8 to the predicted class.
Second, the prediction made by the random forest is not correct.
For each highlighted datapoint we apply SHAP and LIME, which gives importance~$i_0$ for feature~$x^0$ and importance~$i_1$ for feature~$x^1$.
Next, we compute~$\max(|i_0|, |i_1|) - \min(|i_0|, |i_1|)$, which is high if the absolute importance of one feature is higher than the other.
The results are summarized in Figure~\ref{fig:lime_shap_boxplot}, in which we can see that both explanation methods give the impression that one feature is much more important than the other.

Suppose that the doctor would investigate one of the highlighted datapoints.
It would get the impression that the model is very confident, because the output score is at least~0.8, while it is actually incorrect.
Additionally, SHAP and LIME define that one feature is more important than the other.
The prediction and explanation combined suggest that the model can be trusted.
PERFEX is a crucial tool in this scenario because it would inform the doctor that classifier accuracy tends to be low for similar datapoints.

Finally, we investigate why SHAP and LIME indicate that one feature is more important than the other.
The classifier decision boundary is shown in Figure~\ref{fig:lime_shap_decision_boundary}.
The highlighted datapoints are located close to the boundary.
We can see that SHAP and LIME attempt to explain the behavior of the classifier locally, and due to the shape of the boundary both features have varying influence on the predictions.
This also confirms our intuition that SHAP and LIME only explain local behaviour of the classifier.

\section{Case Study: Modality Choices in Mobility}
We present a case study in which we apply PERFEX in the context of mobility.
Cities are facing a transition from conventional mobility concepts such as cars and bikes to so-called new mobility concepts such as ride sharing and e-scooters~\cite{Schade2014}.
To support this transition, policy makers would like to predict and understand existing modality choices for trips in their city.
They use a decision support system which uses a classifier to predict the modality that an individual chooses for a trip, based on trip properties as well as personal characteristics.
The classes correspond to the modalities: car, car as a passenger, public transport, bike, walk.
Each datapoint is a trip consisting of trip properties and characteristics of the traveler.
The trip properties define the travel time for each modality, the cost for car and the cost for public transport.
For the traveler a datapoint defines whether the traveler has a driving license, whether they own a car, and whether they are the main user of the car.
Our dataset consists of 40266 trips from a travel survey conducted by Statistics Netherlands~\cite{CBS2019}.
PERFEX is model-agnostic and applies to any base classifier, but for this specific case study we choose a random forest to illustrate the explanations.
For prediction we train a random forest with 100 trees in the forest and at least 5 datapoints in each leaf.
The accuracy of the final model is 0.91.
We illustrate PERFEX based on two user questions.

\begin{userquestion}
When is the model not able to predict public transport trips as such?
\end{userquestion}
For a mobility researcher analyzing the use of public transport it is important to know whether the model is actually able to label public transport trips as such.
This information can be provided to the researcher by applying our method with the recall of public transport as a performance metric.
\begin{Verbatim}[fontsize=\footnotesize]
There are 4163 trips for which the following
conditions hold:
 travel time public transport > 1800 seconds
 cost public transport > 0.74 euro
 travel time bike <= 1809 seconds
and for these trips the class recall is 0.07
\end{Verbatim}

\begin{userquestion}
When does the model assign high scores to both public transport and bike?
\end{userquestion}
Finally, we consider a mobility researcher that wants to investigate for which trips the model expects that both public transport and bike can be chosen.
In order to answer this question we use a custom performance metric during tree construction.
For each datapoint we take the minimum of the predicted scores for public transport and bike, and the metric~$\mathcal{M}$ takes the mean of these values.
The mean becomes high if the model assigns a high score to both classes.
The explanation below intuitively makes sense: if walking takes a long time and if the traveler does not have a car, then both public transport and biking may be suitable choices.
\begin{Verbatim}[fontsize=\footnotesize]
There are 100 trips for which the following
conditions hold:
 cost public transport <= 21.78 euro
 traveler does not own a car
 travel time walk > 6069 seconds
and for these trips the model assigns on
average at least score 0.19 to both classes
\end{Verbatim}

\section{Conclusions}
We presented PERFEX, a model-agnostic method to create explanations about the performance of a given base classifier.
Our method creates a clustering of a dataset based on a tree structure, such that subsets of data can be distinguished in which a given prediction metric is low or high.
PERFEX can be used to e.g. explain under which circumstances predictions of a model are not accurate, which is highly relevant in the context of building trustworthy decision support systems.
Our experiments have shown that PERFEX can be used to create explanations for various datasets and classification models, even if the base classifier hardly differentiates classes.
It also shows that PERFEX is an important tool in scenarios in which SHAP and LIME are not sufficient to gain trust.
PERFEX currently only uses subsets defined by AND-clauses, and therefore we aim to also investigate other types of subsets in future work~\cite{Speakman2016}.

\section{Acknowledgments}
We received funding from the TNO Appl.AI program, the province of North Brabant in the Netherlands and the SmartwayZ.NL program. We also want to thank Taoufik Bakri and Bachtijar Ashari for preparing our mobility datasets.

\bibliography{bibfile.bib}

\begin{thebibliography}{22}
\providecommand{\natexlab}[1]{#1}

\bibitem[{Antor{\'a}n et~al.(2021)Antor{\'a}n, Bhatt, Adel, Weller, and
  Hern{\'a}ndez-Lobato}]{Antoran2021}
Antor{\'a}n, J.; Bhatt, U.; Adel, T.; Weller, A.; and Hern{\'a}ndez-Lobato,
  J.~M. 2021.
\newblock {Getting a CLUE: A Method for Explaining Uncertainty Estimates}.
\newblock In \emph{Proceedings of the International Conference on Learning
  Representations}.

\bibitem[{Arroyo et~al.(2019)Arroyo, Corea, Jimenez-Diaz, and
  Recio-Garcia}]{Arroyo2019}
Arroyo, J.; Corea, F.; Jimenez-Diaz, G.; and Recio-Garcia, J.~A. 2019.
\newblock {Assessment of Machine Learning Performance for Decision Support in
  Venture Capital Investments}.
\newblock \emph{IEEE Access}, 7: 124233--124243.

\bibitem[{Bertsimas, Orfanoudaki, and Wiberg(2021)}]{Bertsimas2021}
Bertsimas, D.; Orfanoudaki, A.; and Wiberg, H. 2021.
\newblock Interpretable clustering: an optimization approach.
\newblock \emph{Machine Learning}, 110(1): 89--138.

\bibitem[{Breiman(2002)}]{Breiman2002}
Breiman, L. 2002.
\newblock Manual on setting up, using, and understanding random forests v3. 1.
\newblock \emph{Statistics Department University of California Berkeley, CA,
  USA}.

\bibitem[{Breiman et~al.(1984)Breiman, Friedman, Olshen, and
  Stone}]{Breiman1984}
Breiman, L.; Friedman, J.~H.; Olshen, R.~A.; and Stone, C.~J. 1984.
\newblock \emph{Classification and regression trees}.
\newblock Routledge.

\bibitem[{Burkart and Huber(2021)}]{Burkart2021}
Burkart, N.; and Huber, M.~F. 2021.
\newblock {A Survey on the Explainability of Supervised Machine Learning}.
\newblock \emph{Journal of Artificial Intelligence Research}, 70: 245--317.

\bibitem[{CBS(2020)}]{CBS2019}
CBS. 2020.
\newblock {Onderweg in Nederland (ODiN) 2019 -- Onderzoeksbeschrijving}.

\bibitem[{Chen and Guestrin(2016)}]{Chen2016}
Chen, T.; and Guestrin, C. 2016.
\newblock {XGBoost}: A Scalable Tree Boosting System.
\newblock In \emph{Proceedings of the 22nd ACM SIGKDD International Conference
  on Knowledge Discovery and Data Mining}, KDD '16, 785--794. New York, NY,
  USA: ACM.
\newblock ISBN 978-1-4503-4232-2.

\bibitem[{Cox(1958)}]{Cox1958}
Cox, D.~R. 1958.
\newblock The regression analysis of binary sequences.
\newblock \emph{Journal of the Royal Statistical Society: Series B
  (Methodological)}, 20(2): 215--232.

\bibitem[{Dong and Li(1999)}]{Dong1999}
Dong, G.; and Li, J. 1999.
\newblock {Efficient Mining of Emerging Patterns: Discovering Trends and
  Differences}.
\newblock In \emph{Proceedings of the ACM SIGKDD International Conference on
  Knowledge Discovery and Data Mining}, 43--52.

\bibitem[{Dua and Graff(2017)}]{uci:2019}
Dua, D.; and Graff, C. 2017.
\newblock {UCI} Machine Learning Repository.

\bibitem[{{EU High-Level Expert Group on AI}(2019)}]{EU2019}
{EU High-Level Expert Group on AI}. 2019.
\newblock {Ethics Guidelines for Trustworthy AI. European Commission}.

\bibitem[{Guo et~al.(2017)Guo, Pleiss, Sun, and Weinberger}]{Guo2017}
Guo, C.; Pleiss, G.; Sun, Y.; and Weinberger, K.~Q. 2017.
\newblock On calibration of modern neural networks.
\newblock In \emph{Proceedings of the International Conference on Machine
  Learning}, 1321--1330. PMLR.

\bibitem[{Hall et~al.(2019)Hall, Gill, Kurka, and Phan}]{H2O}
Hall, P.; Gill, N.; Kurka, M.; and Phan, W. 2019.
\newblock {Machine Learning Interpretability with H2O Driverless AI}.

\bibitem[{Lundberg and Lee(2017)}]{Lundberg2017}
Lundberg, S.~M.; and Lee, S.-I. 2017.
\newblock A unified approach to interpreting model predictions.
\newblock In \emph{Proceedings of the International Conference on Neural
  Information Processing Systems}, 4768--4777.

\bibitem[{Pan(2002)}]{Pan2002}
Pan, W. 2002.
\newblock Approximate confidence intervals for one proportion and difference of
  two proportions.
\newblock \emph{Computational statistics \& data analysis}, 40(1): 143--157.

\bibitem[{Papanastasopoulos et~al.(2020)Papanastasopoulos, Samala, Chan,
  Hadjiiski, Paramagul, Helvie, and Neal}]{Papanastasopoulos2020}
Papanastasopoulos, Z.; Samala, R.~K.; Chan, H.-P.; Hadjiiski, L.; Paramagul,
  C.; Helvie, M.~A.; and Neal, C.~H. 2020.
\newblock {Explainable AI for medical imaging: deep-learning CNN ensemble for
  classification of estrogen receptor status from breast MRI}.
\newblock In \emph{Medical Imaging 2020: Computer-Aided Diagnosis}, volume
  11314, 228 -- 235. International Society for Optics and Photonics, SPIE.

\bibitem[{Ribeiro, Singh, and Guestrin(2016)}]{Ribeiro2016}
Ribeiro, M.~T.; Singh, S.; and Guestrin, C. 2016.
\newblock {"Why Should I Trust You?" Explaining the Predictions of Any
  Classifier}.
\newblock In \emph{Proceedings of the ACM SIGKDD International Conference on
  Knowledge Discovery and Data Mining}, 1135--1144.

\bibitem[{Ribeiro, Singh, and Guestrin(2018)}]{Ribeiro2018}
Ribeiro, M.~T.; Singh, S.; and Guestrin, C. 2018.
\newblock Anchors: High-precision model-agnostic explanations.
\newblock In \emph{Proceedings of the AAAI Conference on Artificial
  Intelligence}, 1527--1535.

\bibitem[{Schade, Krail, and K{\"u}hn(2014)}]{Schade2014}
Schade, W.; Krail, M.; and K{\"u}hn, A. 2014.
\newblock New mobility concepts: myth or emerging reality?
\newblock In \emph{{Transport Research Arena (TRA)}}.

\bibitem[{Speakman et~al.(2016)Speakman, Somanchi, McFowland~III, and
  Neill}]{Speakman2016}
Speakman, S.; Somanchi, S.; McFowland~III, E.; and Neill, D.~B. 2016.
\newblock Penalized fast subset scanning.
\newblock \emph{Journal of Computational and Graphical Statistics}, 25(2):
  382--404.

\bibitem[{van~der Waa et~al.(2020)van~der Waa, Schoonderwoerd, van Diggelen,
  and Neerincx}]{VanderWaa2020}
van~der Waa, J.; Schoonderwoerd, T.; van Diggelen, J.; and Neerincx, M. 2020.
\newblock Interpretable confidence measures for decision support systems.
\newblock \emph{International Journal of Human-Computer Studies}, 144: 102493.

\end{thebibliography}


\newpage

\noindent\textbf{PERFEX: Classifier Performance Explanations for Trustworthy AI Systems -- Supplementary Material}\\
Erwin Walraven, Ajaya Adhikari, Cor Veenman\\$ $ \\

In this supplement we provide additional results from experiments, details on the experimental setup, and information about the availability of our source code.

\section{Evaluation Results for Other Metrics}
The PERFEX method can be applied to any performance metric~$\mathcal{M}$. In the paper we included only the results of the experiments conducted with accuracy. Table~\ref{table:precision}, \ref{table:recall} and \ref{table:f1} show the results for precision, recall and f1-score (in all cases weighted by support), respectively. This shows that PERFEX can also provide explanations about model performance for other performance metrics.

\section{Details on Experimental Setup}
In our experiments involving multiple datasets and classifiers we use the python library scikit-learn 0.24.2 to train the models with default parameters. For PERFEX we set the minimum number of datapoints per leaf (i.e., $\alpha$) to 100 and the maximum tree depth is 6. Furthermore, we do not split a dataset~$\mathcal{X}$ if the metric value difference after the split is smaller than $0.05$.
In the case study we use the same settings, except that the tree depth is set to 3. Table~\ref{table:datasets} shows for each dataset the number of datapoints, the number of features and the number of classes.

\begin{table}[h]
\begin{tabular}{llll}
\toprule
Dataset & \#datapoints & \#features & \#classes\\
\midrule
Abalone & 4177 & 8 & 3\\
Car evaluation & 1728 & 21 & 4 \\
Contraceptive m.c. & 1473 & 24 & 3 \\
Occ. detection & 10000 & 2 & 2 \\
Gauss. blobs & 10000 & 2 & 3\\
Case study data & 40266 & 11 & 5\\
\bottomrule
\end{tabular}
\caption{\label{table:datasets}Details about datasets used in experiments and case study}
\end{table}

\section{Source code of PERFEX}
The source code of PERFEX can be found as a python package on \url{https://github.com/erwinwalraven/perfex}.

\begin{table*}[t]
\small
\centering
\def\arraystretch{0.8}
\begin{tabular}{llrrrrrrr}
\toprule
\textbf{Dataset}            & $\mathcal{C}$ & \multicolumn{1}{l}{\textbf{Precision of} $\mathcal{C}$} & \multicolumn{1}{l}{\textbf{Num. leafs}} & \multicolumn{1}{l}{\textbf{Tree depth}} & \multicolumn{1}{l}{\textbf{Min precision}} & \multicolumn{1}{l}{\textbf{Max precision}} & \multicolumn{1}{l}{\textbf{MAE}} & \multicolumn{1}{l}{\textbf{STD AE}}  \\
\midrule
\multirow{5}{*}{Abalone}    & SVM        & 38,6\%                                      & 9                                       & 6                                       & 11,1\%                                     & 80,9\%                                     & 8,9\%                            & 5,8\%                                \\
                            & LR         & 56,5\%                                      & 8                                       & 6                                       & 25,0\%                                     & 83,7\%                                     & 14,1\%                           & 14,7\%                               \\
                            & RF         & 52,3\%                                      & 6                                       & 6                                       & 33,9\%                                     & 75,3\%                                     & 4,4\%                            & 1,7\%                                \\
                            & DT         & 51,4\%                                      & 6                                       & 6                                       & 41,1\%                                     & 78,4\%                                     & 2,5\%                            & 2,0\%                                \\
                            & KNN        & 49,4\%                                      & 7                                       & 6                                       & 27,1\%                                     & 81,2\%                                     & 4,4\%                            & 3,6\%                                \\
\midrule
        & SVM        & 95,3\%                                      & 3                                       & 3                                       & 84,7\%                                     & 99,6\%                                     & 3,3\%                            & 2,9\%                                \\
Car                            & LR         & 90,6\%                                      & 4                                       & 4                                       & 73,6\%                                     & 100,0\%                                    & 2,2\%                            & 2,9\%                                \\
evaluation                            & RF         & 95,7\%                                      & 2                                       & 2                                       & 88,8\%                                     & 99,7\%                                     & 1,8\%                            & 0,1\%                                \\
                            & DT         & 96,2\%                                      & 2                                       & 2                                       & 91,6\%                                     & 99,6\%                                     & 2,2\%                            & 0,8\%                                \\
                            & KNN        & 86,9\%                                      & 4                                       & 4                                       & 67,2\%                                     & 100,0\%                                    & 3,4\%                            & 3,6\%                                \\
\midrule
   & SVM        & 34,0\%                                      & 3                                       & 3                                       & 20,3\%                                     & 63,6\%                                     & 15,2\%                           & 15,2\%                               \\
Contraceptive                            & LR         & 52,7\%                                      & 4                                       & 4                                       & 41,5\%                                     & 64,7\%                                     & 7,5\%                            & 5,1\%                                \\
method                            & RF         & 52,3\%                                      & 3                                       & 3                                       & 42,0\%                                     & 61,1\%                                     & 6,8\%                            & 5,7\%                                \\
choice                            & DT         & 52,4\%                                      & 3                                       & 3                                       & 44,1\%                                     & 64,0\%                                     & 3,6\%                            & 3,4\%                                \\
                            & KNN        & 47,6\%                                      & 4                                       & 3                                       & 35,0\%                                     & 65,4\%                                     & 7,1\%                            & 4,8\%                                \\
\midrule
  & SVM        & 86,5\%                                      & 13                                      & 6                                       & 26,3\%                                     & 99,5\%                                     & 3,8\%                            & 3,4\%                                \\
Occupancy                            & LR         & 80,8\%                                      & 7                                       & 6                                       & 9,1\%                                      & 97,3\%                                     & 4,0\%                            & 3,6\%                                \\
detection                            & RF         & 94,4\%                                      & 8                                       & 6                                       & 84,2\%                                     & 99,1\%                                     & 3,3\%                            & 2,7\%                                \\
                            & DT         & 93,5\%                                      & 9                                       & 6                                       & 72,6\%                                     & 98,4\%                                     & 3,9\%                            & 2,5\%                                \\
                            & KNN        & 88,3\%                                      & 12                                      & 6                                       & 66,1\%                                     & 99,5\%                                     & 3,9\%                            & 2,5\%                                \\
\midrule
 & SVM        & 80,5\%                                      & 10                                      & 6                                       & 70,4\%                                     & 100,0\%                                    & 5,8\%                            & 5,9\%                                \\
Gaussian                            & LR         & 80,3\%                                      & 7                                       & 6                                       & 72,7\%                                     & 100,0\%                                    & 4,9\%                            & 4,7\%                                \\
blobs                            & RF         & 78,0\%                                      & 8                                       & 6                                       & 71,6\%                                     & 100,0\%                                    & 3,5\%                            & 2,9\%                                \\
                            & DT         & 77,6\%                                      & 6                                       & 6                                       & 71,7\%                                     & 95,2\%                                     & 3,5\%                            & 2,9\%                                \\
                            & KNN        & 75,4\%                                      & 7                                       & 6                                       & 65,4\%                                     & 99,1\%                                     & 2,3\%                            & 2,0\%                               \\
\bottomrule
\end{tabular}
\caption{\label{table:precision}Evaluation of PERFEX using several datasets and base classifiers with precision as metric~$\mathcal{M}$}
\end{table*}

$ $ \\
\newpage
$ $ \\

\begin{table*}[t]
\small
\centering
\def\arraystretch{0.8}
\begin{tabular}{llrrrrrrr}
\toprule
\textbf{Dataset}            & $\mathcal{C}$ & \multicolumn{1}{l}{\textbf{Recall of }$\mathcal{C}$} & \multicolumn{1}{l}{\textbf{Num. leafs}} & \multicolumn{1}{l}{\textbf{Tree depth}} & \multicolumn{1}{l}{\textbf{Min recall}} & \multicolumn{1}{l}{\textbf{Max recall}} & \multicolumn{1}{l}{\textbf{MAE}} & \multicolumn{1}{l}{\textbf{STD AE}}  \\
\midrule
\multirow{5}{*}{Abalone}    & SVM        & 53,6\%                                   & 6                                       & 6                                       & 33,1\%                                  & 84,3\%                                  & 5,8\%                            & 4,6\%                                \\
                            & LR         & 56,5\%                                   & 6                                       & 6                                       & 36,5\%                                  & 83,5\%                                  & 5,2\%                            & 4,2\%                                \\
                            & RF         & 54,1\%                                   & 7                                       & 6                                       & 29,4\%                                  & 81,9\%                                  & 5,1\%                            & 2,9\%                                \\
                            & DT         & 53,4\%                                   & 6                                       & 6                                       & 42,3\%                                  & 79,5\%                                  & 4,8\%                            & 2,6\%                                \\
                            & KNN        & 49,0\%                                   & 7                                       & 6                                       & 25,2\%                                  & 80,3\%                                  & 4,5\%                            & 5,7\%                                \\
\midrule
        & SVM        & 95,0\%                                   & 2                                       & 2                                       & 87,6\%                                  & 99,1\%                                  & 1,1\%                            & 0,7\%                                \\
Car                            & LR         & 90,5\%                                   & 4                                       & 4                                       & 74,8\%                                  & 100,0\%                                 & 1,9\%                            & 2,9\%                                \\
evaluation                            & RF         & 94,8\%                                   & 2                                       & 2                                       & 88,6\%                                  & 98,2\%                                  & 2,2\%                            & 1,0\%                                \\
                            & DT         & 96,3\%                                   & 2                                       & 2                                       & 90,8\%                                  & 99,4\%                                  & 1,3\%                            & 0,3\%                                \\
                            & KNN        & 87,5\%                                   & 4                                       & 4                                       & 65,5\%                                  & 99,4\%                                  & 3,8\%                            & 3,9\%                                \\
\midrule
    & SVM        & 44,3\%                                   & 3                                       & 3                                       & 33,0\%                                  & 63,3\%                                  & 7,2\%                            & 5,3\%                                \\
Contraceptive                            & LR         & 53,4\%                                   & 3                                       & 3                                       & 41,1\%                                  & 63,8\%                                  & 8,8\%                            & 2,0\%                                \\
method                            & RF         & 52,7\%                                   & 3                                       & 3                                       & 40,8\%                                  & 63,8\%                                  & 8,2\%                            & 6,8\%                                \\
choice                            & DT         & 51,6\%                                   & 3                                       & 3                                       & 38,7\%                                  & 60,4\%                                  & 3,0\%                            & 0,5\%                                \\
                            & KNN        & 48,2\%                                   & 4                                       & 3                                       & 36,0\%                                  & 62,4\%                                  & 9,5\%                            & 6,2\%                                \\
\midrule
  & SVM        & 86,8\%                                   & 10                                      & 6                                       & 31,1\%                                  & 98,3\%                                  & 3,4\%                            & 2,3\%                                \\
Occupancy                            & LR         & 81,9\%                                   & 10                                      & 6                                       & 24,9\%                                  & 98,2\%                                  & 4,1\%                            & 2,9\%                                \\
detection                            & RF         & 94,6\%                                   & 9                                       & 6                                       & 81,1\%                                  & 99,1\%                                  & 2,6\%                            & 2,0\%                                \\
                            & DT         & 93,5\%                                   & 9                                       & 6                                       & 72,8\%                                  & 97,8\%                                  & 3,4\%                            & 1,9\%                                \\
                            & KNN        & 88,6\%                                   & 8                                       & 6                                       & 64,5\%                                  & 98,6\%                                  & 3,5\%                            & 2,1\%                                \\
\midrule
 & SVM        & 80,3\%                                   & 7                                       & 6                                       & 74,0\%                                  & 99,5\%                                  & 1,9\%                            & 1,9\%                                \\
Gaussian                            & LR         & 80,2\%                                   & 9                                       & 6                                       & 71,1\%                                  & 100,0\%                                 & 2,3\%                            & 2,6\%                                \\
blobs                            & RF         & 78,5\%                                   & 7                                       & 6                                       & 72,3\%                                  & 99,2\%                                  & 2,0\%                            & 3,3\%                                \\
                            & DT         & 77,2\%                                   & 8                                       & 6                                       & 69,0\%                                  & 100,0\%                                 & 2,8\%                            & 2,1\%                                \\
                            & KNN        & 75,6\%                                   & 9                                       & 6                                       & 66,1\%                                  & 99,5\%                                  & 2,3\%                            & 2,1\%                                \\
\bottomrule
\end{tabular}
\caption{\label{table:recall}Evaluation of PERFEX using several datasets and base classifiers with recall as metric~$\mathcal{M}$}
\end{table*}

$ $ \\
\newpage
$ $ \\

\begin{table*}[t]
\small
\centering
\def\arraystretch{0.8}
\begin{tabular}{lllrrrrrrr}
\toprule
\textbf{Dataset}            & $\mathcal{C}$ & \textbf{Split metric} & \multicolumn{1}{l}{\textbf{F1-score of} $\mathcal{C}$} & \multicolumn{1}{l}{\textbf{Num. leafs}} & \multicolumn{1}{l}{\textbf{Tree depth}} & \multicolumn{1}{l}{\textbf{Min f1-score}} & \multicolumn{1}{l}{\textbf{Max f1-score}} & \multicolumn{1}{l}{\textbf{MAE}} & \multicolumn{1}{l}{\textbf{STD AE}}  \\
\midrule
\multirow{5}{*}{Abalone}    & SVM        & f1-score              & 44,3\%                                     & 9                                       & 6                                       & 16,1\%                                    & 77,1\%                                    & 6,5\%                            & 5,9\%                                \\
                            & LR         & f1-score              & 53,5\%                                     & 6                                       & 6                                       & 35,3\%                                    & 77,0\%                                    & 7,9\%                            & 5,7\%                                \\
                            & RF         & f1-score              & 52,8\%                                     & 6                                       & 6                                       & 33,7\%                                    & 75,5\%                                    & 5,2\%                            & 2,2\%                                \\
                            & DT         & f1-score              & 51,3\%                                     & 7                                       & 6                                       & 18,7\%                                    & 77,7\%                                    & 5,4\%                            & 2,7\%                                \\
                            & KNN        & f1-score              & 49,1\%                                     & 6                                       & 6                                       & 31,9\%                                    & 80,9\%                                    & 4,3\%                            & 2,9\%                                \\
\midrule
        & SVM        & f1-score              & 94,9\%                                     & 2                                       & 2                                       & 87,4\%                                    & 99,1\%                                    & 1,2\%                            & 0,6\%                                \\
 Car                           & LR         & f1-score              & 90,2\%                                     & 4                                       & 4                                       & 73,2\%                                    & 100,0\%                                   & 2,2\%                            & 2,9\%                                \\
 evaluation                           & RF         & f1-score              & 95,2\%                                     & 2                                       & 2                                       & 89,3\%                                    & 98,4\%                                    & 1,9\%                            & 1,2\%                                \\
                            & DT         & f1-score              & 96,6\%                                     & 2                                       & 2                                       & 91,3\%                                    & 99,4\%                                    & 2,3\%                            & 1,2\%                                \\
                            & KNN        & f1-score              & 86,9\%                                     & 4                                       & 4                                       & 65,0\%                                    & 99,7\%                                    & 3,4\%                            & 3,6\%                                \\
\midrule
   & SVM        & f1-score              & 35,2\%                                     & 3                                       & 3                                       & 20,1\%                                    & 59,6\%                                    & 6,3\%                            & 6,5\%                                \\
Contraceptive                            & LR         & f1-score              & 52,7\%                                     & 3                                       & 3                                       & 40,0\%                                    & 62,4\%                                    & 8,3\%                            & 3,1\%                                \\
method                            & RF         & f1-score              & 51,1\%                                     & 3                                       & 3                                       & 39,4\%                                    & 60,4\%                                    & 4,0\%                            & 3,1\%                                \\
choice                            & DT         & f1-score              & 52,0\%                                     & 3                                       & 3                                       & 40,5\%                                    & 65,8\%                                    & 3,7\%                            & 1,9\%                                \\
                            & KNN        & f1-score              & 47,7\%                                     & 4                                       & 3                                       & 33,3\%                                    & 63,7\%                                    & 8,2\%                            & 6,2\%                                \\
\midrule
  & SVM        & f1-score              & 85,6\%                                     & 7                                       & 6                                       & 44,9\%                                    & 97,9\%                                    & 4,2\%                            & 3,6\%                                \\
 Occupancy                           & LR         & f1-score              & 81,1\%                                     & 8                                       & 6                                       & 15,5\%                                    & 98,1\%                                    & 4,6\%                            & 3,7\%                                \\
  detection                          & RF         & f1-score              & 94,5\%                                     & 9                                       & 6                                       & 74,1\%                                    & 100,0\%                                   & 2,4\%                            & 2,2\%                                \\
                            & DT         & f1-score              & 93,5\%                                     & 9                                       & 6                                       & 76,6\%                                    & 98,3\%                                    & 4,8\%                            & 2,7\%                                \\
                            & KNN        & f1-score              & 88,4\%                                     & 10                                      & 6                                       & 64,5\%                                    & 98,9\%                                    & 4,1\%                            & 3,5\%                                \\
\midrule
 & SVM        & f1-score              & 80,4\%                                     & 8                                       & 6                                       & 74,2\%                                    & 100,0\%                                   & 3,0\%                            & 4,4\%                                \\
 Gaussian                           & LR         & f1-score              & 80,2\%                                     & 9                                       & 6                                       & 73,1\%                                    & 100,0\%                                   & 3,7\%                            & 5,2\%                                \\
   blobs                         & RF         & f1-score              & 78,3\%                                     & 8                                       & 6                                       & 70,0\%                                    & 98,9\%                                    & 2,8\%                            & 2,3\%                                \\
                            & DT         & f1-score              & 77,3\%                                     & 8                                       & 6                                       & 69,7\%                                    & 98,7\%                                    & 4,2\%                            & 3,2\%                                \\
                            & KNN        & f1-score              & 75,5\%                                     & 8                                       & 6                                       & 68,5\%                                    & 100,0\%                                   & 4,9\%                            & 2,9\%                               \\
\bottomrule
\end{tabular}
\caption{\label{table:f1}Evaluation of PERFEX using several datasets and base classifiers with f1-score as metric~$\mathcal{M}$}
\end{table*}

\end{document}